\documentclass[letterpaper]{article} 
\usepackage[submission]{aaai24}  
\usepackage{times}  
\usepackage{helvet}  
\usepackage{courier}  
\usepackage[hyphens]{url}  
\usepackage{graphicx} 
\usepackage{natbib}  
\usepackage{caption} 
\usepackage{algorithm}
\usepackage{algorithmic}

\usepackage{newfloat}
\usepackage{listings}
\DeclareCaptionStyle{ruled}{labelfont=normalfont,labelsep=colon,strut=off} 
\floatstyle{ruled}
\newfloat{listing}{tb}{lst}{}
\floatname{listing}{Listing}
\usepackage{epsfig}
\usepackage{amsmath}
\usepackage{amssymb}
\usepackage{booktabs}
\usepackage{multirow}
\usepackage{eso-pic}
\usepackage{tikz}
\usepackage{everypage}
\usepackage{float}
\usepackage{placeins}
\usepackage[switch]{lineno}

\title{Towards Omni-supervised Referring Expression Segmentation}
\author{
    Written by AAAI Press Staff\textsuperscript{\rm 1}\thanks{With help from the AAAI Publications Committee.}\\
    AAAI Style Contributions by Pater Patel Schneider,
    Sunil Issar,\\
    J. Scott Penberthy,
    George Ferguson,
    Hans Guesgen,
    Francisco Cruz\equalcontrib,
    Marc Pujol-Gonzalez\equalcontrib
}
\affiliations{
    \textsuperscript{\rm 1}Association for the Advancement of Artificial Intelligence\\

    1900 Embarcadero Road, Suite 101\\
    Palo Alto, California 94303-3310 USA\\
    proceedings-questions@aaai.org
}

\usepackage{bibentry}

\begin{document}

\maketitle

\begin{abstract}
Referring Expression Segmentation (RES) is an emerging task in computer vision, which segments the target instances in images based on text descriptions. However, its development is plagued by the expensive segmentation labels. 
To address this issue, we propose a new learning task for RES called~\emph{Omni-supervised Referring Expression Segmentation} (Omni-RES), which aims to make full use of unlabeled, fully labeled and weakly labeled data,~\emph{e.g.}, referring points or grounding boxes, for efficient RES training.
To accomplish this task, we also propose a novel yet strong baseline method for Omni-RES based on the recently popular teacher-student learning, where the weak labels are not directly transformed into supervision signals but used as a yardstick to select and refine high-quality pseudo-masks for teacher-student learning.
To validate the proposed Omni-RES method, we apply it to a set of state-of-the-art RES models and conduct extensive experiments on a bunch of RES datasets. The experimental results yield the obvious merits of Omni-RES than the fully-supervised and semi-supervised training schemes. For instance, with only 10\% fully labeled data, Omni-RES can help the base model achieve 100\% fully supervised performance, and it also outperform the semi-supervised alternative by a large margin,~\emph{e.g.}, +14.93\% on RefCOCO and +14.95\% on RefCOCO+, respectively.
More importantly, Omni-RES also enable the use of large-scale vision-langauges like Visual Genome to facilitate low-cost RES training, and achieve new SOTA performance of RES, \emph{e.g.}, 80.66 on RefCOCO.
Our code is anonymously released at: \url{https://github.com/nineblu/omni-res}.
\end{abstract}

\begin{figure}[!htb]
\centering
\includegraphics[width=0.44\textwidth]{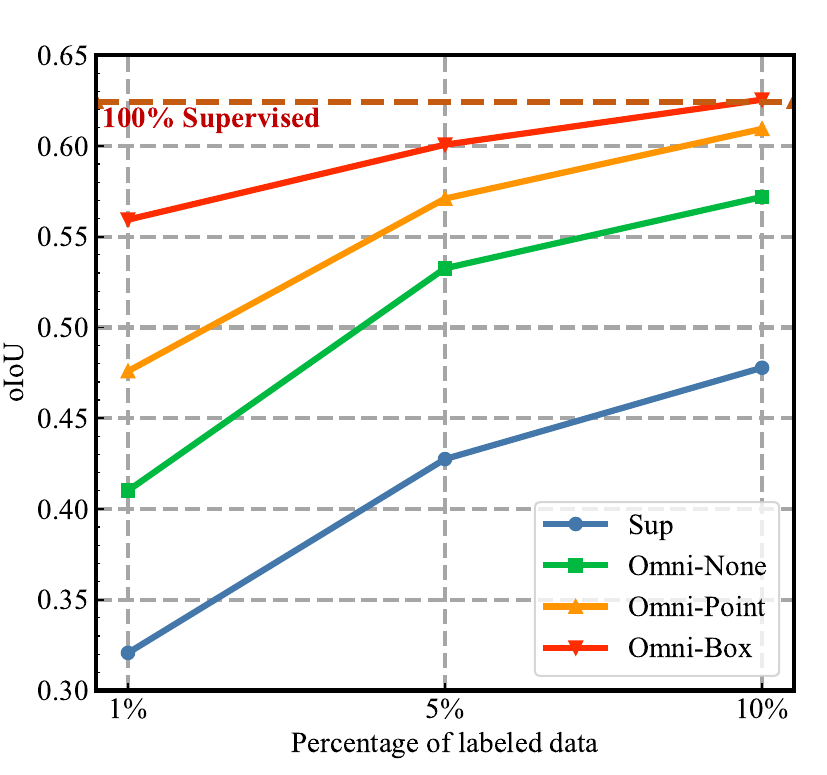}
\caption{Omni-RES significantly improve the performance of RES with limited mask annotations. With 10\% fully labeled data, Omni-RES can help our base model SimRES achieve 100\% supervised performance.}
\label{fig:ref_iou}
\end{figure}

\section{Introduction}

Referring Expression Segmentation (RES)~\cite{RefCOCO,MCN,CGAN,SeqTR}, also called~\emph{Referring Image Segmentation}~\cite{LTS,MaIL,CRIS}, aims to ground the visual instances via binary segmentation masks according to the given natural language expressions.
Compared with conventional segmentation tasks~\cite{FCN, Mask_RCNN}, RES merits in its flexibility and generalization ability, where the segmentation can be flexibly decided by the text descriptions. Thus, it has great potential in a wide range of applications, such as video surveillance, image editing and human-computer interaction. However, its annotation, \emph{i.e.}, segmentation masks, is time-consuming and labor-intensive, which greatly plagues the its development.

However, there exists massive high-quality but weakly-labeled data in vision-language research~\cite{SBU,COCO,VG,CC}.
For instance, the popular Visual Genome dataset~\cite{VG} has over 6 millions of grounding boxes, which are also annotated according to the text expression. More recently, these additional data has been exploited to boost the development of referring expression comprehension (REC) \cite{SimREC,SeqTR}, a similar task to RES. But so far, there is still no feasible way for RES to exploit these much cheaper annotations.

Inspired by the recent progress in computer vision~\cite{Omni-DETR}, we propose a new training task for RES towards the effective use of low-cost VL labels, termed \emph{Omni-supervised Referring Expression Segmentation} (Omni-RES). 
Compared with the semi-supervised or weakly-supervised learning schemes~\cite{https://doi.org/10.48550/arxiv.2205.04725,https://doi.org/10.48550/arxiv.2212.10278,9875225}, Omni-RES has no strict restrictions on its training data, which can be the unlabeled, fully labeled or weakly labeled ones. 
In particular, we focus on the exploration of the weak labels of grounding boxes and referring points, which are much cheaper to annotate and easier to obtain from existing datasets. By including these weakly-labeled data, we can achieve much better performance than fully-supervised learning, which is even on par with the fully supervised performance at the same data scale, as shown in Fig.~\ref{fig:ref_iou}.

\begin{figure*}
\begin{center}
\includegraphics[scale=0.64]{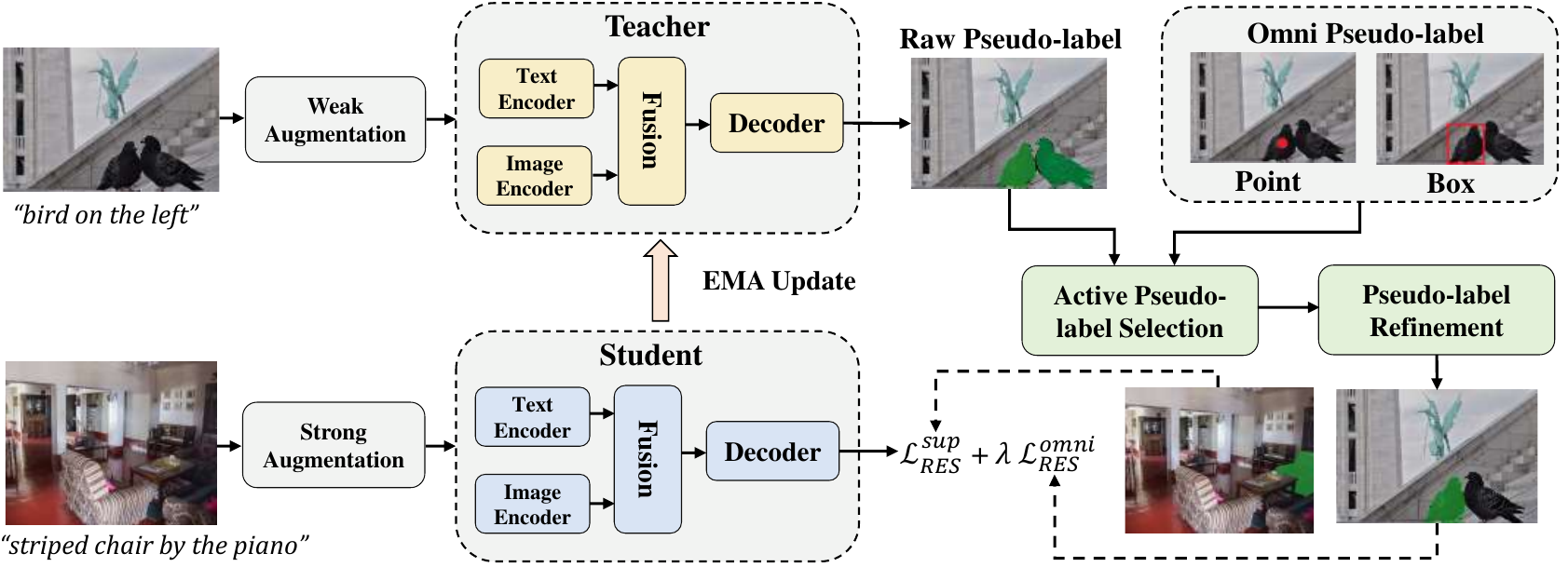}
\end{center}
   \caption{\textbf{Overall of the proposed baseline training paradigm for Omni-RES.} This paradigm uses a teacher-student framework, where the Teacher and the Student networks are initialized with the same RES models. During omni-training, the teacher is in charged of producing pseudo segmentation masks for the student with the limited fully label information. The omni-labels,~\emph{i.e.}, point or box, are applied to actively refine the raw pseudo-labels, thereby improving the teacher-student learning.}
\label{fig:omni_res}
\end{figure*}

To accomplish Omni-RES, we further propose a strong baseline method in this paper, of which structure is depicted in Fig.~\ref{fig:omni_res}.
This baseline scheme applies the teacher-student learning~\cite{UnbiasedTeacher,Active_Teacher} to train the RES models via the self-consistency based objectives with limited label information. 
In terms of the weak labels, \emph{e.g.}, grounding boxes or points, they are not directly transformed into supervision signals through complex operations, as most existing weakly supervised methods do in other tasks~\cite{https://doi.org/10.48550/arxiv.1803.10464,https://doi.org/10.48550/arxiv.2009.12547,https://doi.org/10.48550/arxiv.2004.04581,https://doi.org/10.48550/arxiv.2007.01947,https://doi.org/10.48550/arxiv.2104.00905}. Instead, we use them as yardsticks to select and produce high-quality pseudo-masks predicted by the teacher. The intuition is that the quality of pseudo-labels plays a critical role in teacher-student learning~\cite{UnbiasedTeacher,SoftTeacher,https://doi.org/10.48550/arxiv.2106.00168}, which is decisive for the final performance.
With the above designs, Omni-RES can well exploit massive unlabeled and low-cost labeled data with limited label information, thereby achieving inexpensive RES training.

To validate Omni-RES, we first propose a base model based on a lightweight REC model~\cite{SimREC} for quick validations, termed SimRES. Meanwhile, we also examine Omni-RES on two receently proposed RES models, namely LAVT~\cite{LAVT} and ReLA~\cite{GRES}. 
Numerous experiments are conducted on three RES benchmarks~\cite{RefCOCO,umd,google}.
The experimental results show that Omni-RES can achieve +24.59\% and +16.24\% improvements on average with omni-labels of grounding boxes and points, respectively.
Notably, Omni-RES can help SimRES achieve 100\% fully supervised performance with only 10\% labeled data.
Its advantages to the semi-supervised baseline are also distinct,~\emph{e.g.}, +14.93\% on RefCOCO with 1\% labeled data.
Meanwhile, Omni-RES also help ReLA achieve new SOTA performance on these benchmarks,~\emph{e.g.}, 80.66 on RefCOCO~\textit{testA}.

Overall, our contribution is three-fold.

\begin{itemize}
\item We introduce a new learning task for RES to address the issue of insufficient label information, termed Omni-RES, which aims to use unlabeled, weakly labeled and fully labeled data for efficient RES training.
\item To accomplish Omni-RES, we also propose a strong baseline method with a novel teacher-student learning scheme, where the weak labels are used as a yardstick to select and refine pseudo-masks.
\item On three benchmark datasets, Omni-RES can achieve superior performance gains than the fully-supervised and semi-supervised baselines. 
Notably, with only 10\% fully labeled data, Omni-RES can help the model achieve 100\% supervised performance.
\end{itemize}

\section{Related Work}

\subsection{Referring Expression Segmentation}

Referring expression segmentation (RES) aims to segment the target object in an image according to a given natural language expression.~\cite{Segmentation_from_Natural_Language_Expressions} Early RES works~\cite{https://doi.org/10.48550/arxiv.1812.03299,MAttNet} usually follow a the two-stage pipeline, and formulate RES as an object-expression matching problem. Recently, one-stage RES models~\cite{Recurrent_Multimodal_Interaction_for_Referring_Image_Segmentation,DMN,RRN} have attracted increasing attentions recently, where the RES model embeds the text features into a segmentation network, and directly predict the mask of the referent. Inspired by the great success of Transformer, a variety of Transformer-based RES models are proposed recently~\cite{VLT,ReSTR,RefTR,vpd,GRES,polyformer}. VLT~\cite{VLT} takes fusion of vision and language as input to transformer encoder-decoder framework, and ReSTR~\cite{ReSTR} uses transformer encoders for the two modalities independently.
In addition to the design of network structure, some recent advances also investigate the early fusion strategy~\cite{MaIL,LAVT} and the multi-modal alignment loss~\cite{CRIS} to boost performance.

\subsection{Omni-Supervised Learning}

Omni-supervised learning (OSL) aims to combine different forms of annotations,~\emph{i.e.,} tag and point, to supervise the model~\cite{https://doi.org/10.48550/arxiv.1712.04440,UFO2,Omni-DETR}. 
~\cite{https://doi.org/10.48550/arxiv.1712.04440} is the first to propose the idea of omni-supervised learning, which uses internet-scale sources of unlabeled data for keypoint detection and object detection. 
After that, UFO$^2$~\cite{UFO2} considers more forms of annotation inferior to box as weakly labeled annotations for object detection, such as tag, point and scribble. 
In this method, tag can provides the information of class about the object which is hoped to be detected in the image. Similarly, point and scribble can provide some level of  localization information of the object in the image, which facilitates the omni-supervised learning, far better than just using unlabeled data.
Recently, Omni-DETR~\cite{Omni-DETR}, based on DETR~\cite{DETR}, proposes a unified framework to combine different forms of annotations, which maximizes the use of weakly labeled data for object detection, including tags, points without tags, boxes without tags, points with tags and tags with counts. Although OSL has been widely applied in object detection, it still lacks an exploration on segmentation-related task~\cite{FCN,DeepLab,Mask_RCNN,BlendMask,Panoptic_Segmentation,https://doi.org/10.48550/arxiv.1808.03575,lavin,laconvnet}. 
In this paper, we commit the first attempt on the segmentation-based task,~\emph{i.e.}, RES, which allows for the integration of points and boxes as weakly labeled annotations.

\section{Method}
\subsection{Task Definition}
Omni-supervised referring expression segmentation (Omni-RES) aims to exploit unlabeled, fully labeled and weakly labeled data to facilitate model training. Given a set of fully labeled data $D_f=\{(x_i,t_i,y_i^f)\}_{i=1}^{N}$ and omni-labeled data $D_o=\{(x_j,t_j,y_j^o)\}_{j=1}^{M}$, the objective of omni-supervised RES  $\mathcal{L}$ is formulated as
\begin{equation}
\label{overall_equ}
    \begin{aligned} 
\min  \mathcal{L}(\theta;D_{f},D_{o}). 
    \end{aligned}
\end{equation}
Here, $x$, $t$ and $y$ denote the input images, texts and labels, respectively. $N$ and $M$ are the numbers of fully-labeled and omni-labeled examples. $\theta$ denotes the model's parameters.

In particular, the effectiveness of omni-supervised learning is highly dependent on the design of omni-labels $y_j^o$~\cite{Omni-DETR}. Omni-labels need to provide useful information for RES, and should be easier to access than the segmentation masks. In this case, we define three types of omni-labels below:
\begin{itemize}
    \item \textbf{None ($y^o=\varnothing$)}. It means no annotations in $D_o$. In this case, Omni-RES can also be regraded as a semi-supervised learning process in this paper, when $D_o$ is the unlabeled data. 

    \item \textbf{Point ($y^o=\{p_j\}_{j=1}^{M}$)}. The grounding point is defined by the geometric center of the referred object, as shown in Fig.\ref{fig:omni_res}. Point tag   is actually cheap and effortless, but can provide the location prior of the referent, which is critical for RES task.

    \item \textbf{Box ($y^o=\{b_j\}_{j=1}^{M}$)}. It denotes the bounding box of the target object. This type of labeling is slightly more expensive than point, but still much easer than the binary mask of RES. As revealed in~\cite{MCN}, grounding box can well help the RES model locate the object and refine the segmentation results.
\end{itemize}

Compared with existing OSL works~\cite{UFO2,Omni-DETR}, we omit other weak labels such as semantic tags, object counts and visual traces (scribbled). To explain, semantic tags are redundant to the text information of RES's expressions, and object counts are of limited importance to RES. The visual traces and scribbles are potentially helpful to Omni-RES, but their access are relatively difficult than box and points in existing VL datasets.

\subsection{Omni-RES}

\subsubsection{Overall framework.}
To achieve Omni-RES, we propose a strong baseline training paradigm in this paper, of which framework is depicted in Fig.~\ref{fig:omni_res}. In particular, the optimization objective of Omni-RES is defined by 
\begin{equation}
\label{omni_equ}
    \begin{aligned} 
\mathcal{L}=\mathcal{L_{\text{sup}}}+ \lambda \mathcal{L_{\text{omni}}}, 
    \end{aligned}
\end{equation}
where $\mathcal{L_{\text{sup}}}$ and $\mathcal{L_{\text{omni}}}$ denote the supervised and omni-supervised objective functions, respectively. $\lambda$ is the hyperparameter to adjust the weight of $L_{\text{omni}}$. For RES, the supervised objective is defined by 
\begin{equation}
\label{sup_equ}
    \begin{aligned} 
\mathcal{L_{\text{sup}}}(\theta;x_i,t_i,y_i^f)=& -[y_i^f\log (G(x_i,t_i|\theta))\\
&+(1-y_i^f)\log (1-G(x_i,t_i|\theta)) ],
    \end{aligned}
\end{equation}
where $G(x_i,t_i|\theta)$ denotes the binary predictions.  

To maximize the benefits of omni-supervised learning, it is essential to design an effective objective $\mathcal{L_{\text{omni}}}$ for RES. In existing segmentation research~\cite{CAM,OAA,SEAM,AffinityNet}, these weak labels are often transformed into direct supervisions via \emph{Classification Activation Map}~\cite{CAM} or \emph{Self-supervised Equivariant Attention Mechanism}~\cite{SEAM}. 
However, RES examples often contains more sparser label information, \emph{e.g.}, only one target corresponding to the expression. And existing weakly-supervised methods still cannot yield supervision signals with satisfactory quality for RES~\cite{https://doi.org/10.48550/arxiv.2205.04725,https://doi.org/10.48550/arxiv.2212.10278,9875225}.
In this case, we regard Omni-RES as a teacher-student based pseudo-learning process.

In particular, Omni-RES consists of two networks with the same configurations, namely teacher and student. Similar to semi-supervised methods~\cite{STAC,UnbiasedTeacher,SoftTeacher}, the teacher is used to predict pseudo-labels for optimizing the student, while the parameters of the student will be used to update the teacher via EMA~\cite{EMA}, as shown in Figure~\ref{fig:omni_res}.

During teacher-student learning, Omni-labels are leveraged to select and refine the pseudo-labels, thereby significantly improving the effectiveness of teacher-student learning. To this end, $\mathcal{L_{\text{omni}}}$ is defined by
\begin{equation}
\label{omni_equ_}
\begin{aligned} 
\mathcal{L}_\text{omni}(\theta;x_j,t_j,y_j^o)=& - [y_j^{o'}\log (G(x_j,t_j|\theta))\\
&+(1-y_j^{o'})\log (1-G(x_j,t_j|\theta))],\\
\text{where}  \quad y_j^{o'}=&r(y_i^o,G_t(x_j,t_j|\theta)).
\end{aligned}
\end{equation}
Here, $G(\cdot)$ and $G_t(\cdot)$ denote the student and teacher, respectively  , and $r(\cdot)$ represents  the active pseudo-label refinement schemes proposed in this paper, which are based on omni-labels. 
During training, the teacher's parameters $\theta_t$ are updated from the student via EMA~\cite{EMA}, which can be formulated by 
\begin{equation}
\label{ts_equ}
    \begin{aligned} 
\theta_{t}^{k}\leftarrow \alpha\theta_{t}^{k-1}+\left (1-\alpha \right )\theta^{k},
    \end{aligned}
\end{equation}
where $k$ is the training step and $\alpha$ denotes the keeping rate. To alleviate error accumulation~\cite{UnbiasedTeacher}, we apply strong data augmentations to the input images of the student and teacher, respectively.

\subsubsection{Active Pseudo-Label Refinement} 
In teacher-student learning, the quality of pseudo-labels is decisive for the final performance~\cite{UnbiasedTeacher,SoftTeacher,RPL}.
For Omni-RES, the problem of low-quality pseudo-labels is much more significant than other tasks, \emph{e.g.}, object detection~\cite{STAC,UnbiasedTeacher,SoftTeacher,HumbleTeachers}. At the initial training stage, the model is prone to producing scattered and erroneous pseudo-masks. 

To overcome this challenge, we propose~\textit{Active Pseudo-Label Refinement} (APLR) to refine and filter pseudo-masks by using omni-labels $y_i^o$ as visual priors to obtain high-quality pseudo-label $y_j^{o'}$ for optimizing omni-supervised objective in Eq.~\ref{omni_equ_}, which is an essential procedure in the framework of Omni-RES.
As depicted in Fig.~\ref{fig:omni_res}, APLR consists of two steps. The first step is ~\textit{Active Pseudo-label Selection}, which select the plausible segmented masks filtered with omni-labels from the raw pseudo-labels. The second step is~\textit{Pseudo-label Refinement}, which means that after selection, we keep the selected masks and discard the other masks to refine raw pseudo-labels for training.

Specifically, in~\textit{Active Pseudo-label Selection} , we propose two selection metrics for the omni-labels of point and box, respectively. 

The selection metric for omni-points $s_p$ is formulated by
\begin{equation} 
\label{equ:sel_point}
    \begin{aligned} 
s_p^j=\mathbb{I}(p_j \cap m_j \not= \varnothing ),
    \end{aligned}
\end{equation}
where $m_j$ denotes the pseudo-masks predicted by the teacher. $\mathbb{I}:X \rightarrow \{0, 1\}$ is the indicator function. Eq.~\ref{equ:sel_point} means that if the point annotation does not fall in any segmented area of the pseudo-mask, \emph{i.e.,} $s_p^j=0$, the pseudo-mask is likely to be negative label and will be skipped during training.
In practice, we apply the topological method in~\cite{topological} to disentangle the multiple predicted segmented areas which are not connected to each other in the pseudo-masks, and then select the plausible segmented areas in which point $p_j$ falls.

For omni-boxes, the selection metric $s_b$ is defined by
\begin{equation} 
\label{equ:sel_box}
    \begin{aligned} 
s_b^j=\mathbb{I}(\frac{ \sum b_j'  m_j}{h_b^j \times w_b^j}>\tau),
    \end{aligned}
\end{equation}
where $h_b^j$ and $ w_b^j$ denote the height and width of box $b_j$, and $\tau$ is the threshold.
When the segmented area is zero,~\emph{i.e.}, $b_j'=0$ means that the predicted mask is incorrect, so we can directly discard this negative pseudo-label. Similarly, when the predicted area is smaller than the threshold, \emph{i.e.}, $s_b=0$, the quality of pseudo-mask is poor and can be ignored.

In~\textit{Pseudo-label Refinement}, to obtain refined pseudo-labels $y_j^{o'}$ from raw pseudo-labels, we filter out the predicted masks where $s^j=0$ based on the two selection metrics and keep the rest.

With the above filtering schemes, we can greatly improve the quality of pseudo-labels, thereby boosting the performance of Omni-RES. 
 
\begin{algorithm}
\caption{Pseudo Code of Omni-RES}
\label{alg1} 
\begin{algorithmic}
\STATE \textbf{Input:} Labeled Dataset $\{\mathcal{X}^0,\mathcal{T}^0,\mathcal{Y}_F^0\}$, Omni-labeled Dataset $\{\mathcal{X}^0,\mathcal{T}^0,\mathcal{Y}_O^0\}$, Maximum Iteration K
\STATE \textbf{Output:} Student Model $M^s$
\end{algorithmic}
\begin{algorithmic}[1]

\STATE \textbf{for all} $i=1,...,K$ \textbf{do}
\STATE\hspace{\algorithmicindent} \textbf{for all} $(x,t,y_f)\in \{\mathcal{X}^0,\mathcal{T}^0,\mathcal{Y}_F^0\}$ \textbf{do}
\STATE\hspace{\algorithmicindent}\hspace{\algorithmicindent} Calculate supervised loss using Student network $M_{i-1}^t$ by Eq.~\ref{sup_equ}
\STATE\hspace{\algorithmicindent} \textbf{for all} $(x,t,y_o)\in \{\mathcal{X}^0,\mathcal{T}^0,\mathcal{Y}_O^0\}$ \textbf{do}
\STATE\hspace{\algorithmicindent}\hspace{\algorithmicindent} Output raw pseudo-labels of omni-labeled data using Teacher network $M_{i-1}^t$
\STATE\hspace{\algorithmicindent}\hspace{\algorithmicindent} Filter raw pseudo-labels into refined pseudo-labels using \textit{Active Pseudo-label Refinement} by Eq.~\ref{equ:sel_point} and Eq.~\ref{equ:sel_box}
\STATE\hspace{\algorithmicindent}\hspace{\algorithmicindent} Calculate omni-supervised loss using refined pseudo-labels by Eq.~\ref{omni_equ_}
\STATE\hspace{\algorithmicindent} \textbf{for all} $(x,t,y_f)\in \{\mathcal{X}^0,\mathcal{T}^0,\mathcal{Y}_F^0\}$ and $(x,t,y_o)\in \{\mathcal{X}^0,\mathcal{T}^0,\mathcal{Y}_O^0\}$ \textbf{do}
\STATE\hspace{\algorithmicindent} Update the parameters of Student $M_i^s$ by Eq.~\ref{omni_equ}
\STATE\hspace{\algorithmicindent} Update the parameters of Teacher $M_i^t$ by Eq.~\ref{ts_equ}
\STATE\hspace{\algorithmicindent} \textbf{end for}
\STATE \textbf{end for}
\STATE \textbf{return} $M_K^s$
\end{algorithmic}
\end{algorithm}


\subsubsection{Base Model}
For a quick validation, we further revise a recent REC model called SimREC~\cite{SimREC} as our base model. As shown in Fig.~\ref{fig:omni_res}, given the input image and text, the visual backbone and the text encoder extract   features of two modalities, upon which the multi-modal fusion branch is used to conduct multi-modal interactions. Finally, the  decoder is applied to predict the mask of the referent.

\section{Experiments}

\subsection{Datasets and Metrics}

\textbf{RefCOCO, RefCOCO+ and RefCOCOg} are three widely-used RES benchmark datasets~\cite{RefCOCO,umd,google}. In particular, RefCOCO and RefCOCO+ contain 140\textit{k} natural language expressions for 50\textit{k} objects in 20\textit{k} images.  The descriptions of RefCOCO are mainly about spatial content, whereas RefCOCO+ has a greater prevalence of attribute and relative spatial descriptions. In RefCOCO and RefCOCO+, their splits \textit{testA} and \textit{testB} are about people and objects, respectively. RefCOCOg has a comparable dataset size to RefCOCO and RefCOCO+, \emph{i.e.,} 104\textit{k} expressions for 54\textit{k} objects in 26\textit{k} image, but its descriptions are more complex.
RefCOCOg has two different partitions,~\emph{i.e.}, google~\cite{google} and UMD partitions~\cite{umd}. In this paper, we use UMD partition for comparison.

\textbf{Visual Genome}~\cite{VG} is a large-scale vision-language dataset, containing 108,077 images taken from the MSCOCO dataset, 1.7M vqa examples, 5.4M region descriptions, 3.8M object instances, 2.8M attributes, 2.3M  visual relationships. Here, the region descriptions and corresponding boxes can be used as the omni-labels.

\textbf{Metrics.} Following the settings of~\cite{LAVT, GRES, polyformer}, We use overall~\emph{intersection-over-union (oIoU)},~\emph{mean intersection-over-union (mIoU)}, and~\emph{precision} as the evaluation metrics for our experiments.

\begin{table*}[t]
\centering
\caption{Ablation study of Omni-RES with different types of omni-labels. Here, Omni-None means that the unlabeled data are used, which can be regarded as the semi-supervise baseline.}
\begin{tabular}{cccccccccc}
\toprule
\multirow{2}{*}{Method} & \multirow{2}{*}{Percentage} & \multicolumn{3}{c}{\centering RefCOCO} & \multicolumn{3}{c}{RefCOCO+} & \multicolumn{2}{c}{RefCOCOg} \\ \cline{3-10} \specialrule{0em}{1pt}{1pt}
& & \multicolumn{1}{c}{val} & \multicolumn{1}{c}{testA} & testB & \multicolumn{1}{c}{val} & \multicolumn{1}{c}{testA} & testB & \multicolumn{1}{c}{val} & test \\ \midrule
\multicolumn{1}{c}{Fully Supervised} & 1\% & \multicolumn{1}{c}{32.07} & \multicolumn{1}{c}{34.79} & \multicolumn{1}{c}{27.61} & \multicolumn{1}{c}{19.45} & \multicolumn{1}{c}{21.60} & \multicolumn{1}{c}{15.11} & \multicolumn{1}{c}{24.53} & 24.05 \\
\multicolumn{1}{c}{Omni-None} & 1\% & \multicolumn{1}{c}{41.01} & \multicolumn{1}{c}{44.31} & \multicolumn{1}{c}{37.55} & \multicolumn{1}{c}{23.46} & \multicolumn{1}{c}{24.03} & \multicolumn{1}{c}{20.27} & \multicolumn{1}{c}{29.69} & 29.38 \\
\multicolumn{1}{c}{Omni-Point} & 1\% & \multicolumn{1}{c}{47.59} & \multicolumn{1}{c}{50.04} & \multicolumn{1}{c}{45.55} & \multicolumn{1}{c}{33.18} & \multicolumn{1}{c}{36.47} & \multicolumn{1}{c}{29.47} & \multicolumn{1}{c}{35.24} & 37.11 \\
\multicolumn{1}{c}{Omni-Box} & 1\% & \multicolumn{1}{c}{55.94} & \multicolumn{1}{c}{59.26} & \multicolumn{1}{c}{53.04} & \multicolumn{1}{c}{41.12} & \multicolumn{1}{c}{42.99} & \multicolumn{1}{c}{35.88} & \multicolumn{1}{c}{43.83} & 46.18 \\ \midrule
\multicolumn{1}{c}{Fully Supervised} & 5\% & \multicolumn{1}{c}{42.75} & \multicolumn{1}{c}{45.26} & \multicolumn{1}{c}{41.53} & \multicolumn{1}{c}{28.18} & \multicolumn{1}{c}{29.01} & \multicolumn{1}{c}{24.71} & \multicolumn{1}{c}{32.73} & 34.13 \\
\multicolumn{1}{c}{Omni-None} & 5\% & \multicolumn{1}{c}{53.26} & \multicolumn{1}{c}{55.82} & \multicolumn{1}{c}{50.70} & \multicolumn{1}{c}{36.55} & \multicolumn{1}{c}{38.96} & \multicolumn{1}{c}{31.21} & \multicolumn{1}{c}{39.68} & 39.25 \\
\multicolumn{1}{c}{Omni-Point} & 5\% & \multicolumn{1}{c}{57.10} & \multicolumn{1}{c}{60.59} & \multicolumn{1}{c}{55.10} & \multicolumn{1}{c}{42.76} & \multicolumn{1}{c}{46.72} & \multicolumn{1}{c}{36.95} & \multicolumn{1}{c}{45.29} & 45.24 \\
\multicolumn{1}{c}{Omni-Box} & 5\% & \multicolumn{1}{c}{60.07} & \multicolumn{1}{c}{62.89} & \multicolumn{1}{c}{58.22} & \multicolumn{1}{c}{46.78} & \multicolumn{1}{c}{49.09} & \multicolumn{1}{c}{38.83} & \multicolumn{1}{c}{48.61} & 49.00 \\ \midrule
\multicolumn{1}{c}{Fully Supervised} & 10\% & \multicolumn{1}{c}{47.78} & \multicolumn{1}{c}{48.62} & \multicolumn{1}{c}{45.58} & \multicolumn{1}{c}{31.69} & \multicolumn{1}{c}{32.47} & \multicolumn{1}{c}{27.84} & \multicolumn{1}{c}{38.39} & 37.57 \\
\multicolumn{1}{c}{Omni-None} & 10\% & \multicolumn{1}{c}{57.19} & \multicolumn{1}{c}{60.04} & \multicolumn{1}{c}{53.97} & \multicolumn{1}{c}{41.48} & \multicolumn{1}{c}{43.41} & \multicolumn{1}{c}{36.22} & \multicolumn{1}{c}{43.90} & 43.44 \\
\multicolumn{1}{c}{Omni-Point} & 10\% & \multicolumn{1}{c}{60.94} & \multicolumn{1}{c}{64.02} & \multicolumn{1}{c}{58.60} & \multicolumn{1}{c}{45.89} & \multicolumn{1}{c}{48.36} & \multicolumn{1}{c}{39.09} & \multicolumn{1}{c}{47.08} & 46.74 \\
\multicolumn{1}{c}{Omni-Box} & 10\% & \multicolumn{1}{c}{62.55} & \multicolumn{1}{c}{64.74} & \multicolumn{1}{c}{58.76} & \multicolumn{1}{c}{47.85} & \multicolumn{1}{c}{50.04} & \multicolumn{1}{c}{41.61} & \multicolumn{1}{c}{49.41} & 49.08 \\
\bottomrule
\end{tabular}
\label{tab2}
\end{table*}

\begin{table*}[t]
\centering
\caption{Comparison of the proposed Active Pseudo-Label Refinement (APLR) with alternative pseudo-label filtering schemes. The proportion of fully labeled data is 1\%.}
\setlength{\tabcolsep}{10pt} 
 {
\begin{tabular}{ccccccccc}
\toprule
\multirow{2}{*}{Method} & \multicolumn{3}{c}{\centering RefCOCO} & \multicolumn{3}{c}{RefCOCO+} & \multicolumn{2}{c}{RefCOCOg} \\ \cline{2-9} \specialrule{0em}{1pt}{1pt}
& \multicolumn{1}{c}{val} & \multicolumn{1}{c}{testA} & testB & \multicolumn{1}{c}{val} & \multicolumn{1}{c}{testA} & testB & \multicolumn{1}{c}{val} & test \\ \midrule
\multicolumn{1}{l}{None} & \multicolumn{1}{c}{41.01} & \multicolumn{1}{c}{44.31} & \multicolumn{1}{c}{37.55} & \multicolumn{1}{c}{23.46} & \multicolumn{1}{c}{24.03} & \multicolumn{1}{c}{20.27} & \multicolumn{1}{c}{29.69} & 29.38 \\ \midrule
\multicolumn{1}{l}{point distance} & \multicolumn{1}{c}{40.90} & \multicolumn{1}{c}{44.24} & \multicolumn{1}{c}{36.90} & \multicolumn{1}{c}{26.29} & \multicolumn{1}{c}{28.04} & \multicolumn{1}{c}{21.38} & \multicolumn{1}{c}{29.09} & 29.42 \\
\multicolumn{1}{l}{$APLR_{w/o filtering}$} & \multicolumn{1}{c}{43.81} & \multicolumn{1}{c}{47.05} & \multicolumn{1}{c}{38.64} & \multicolumn{1}{c}{25.44} & \multicolumn{1}{c}{24.80} & \multicolumn{1}{c}{19.54} & \multicolumn{1}{c}{31.40} & 31.09 \\
\multicolumn{1}{l}{$APLR_{point}$} & \multicolumn{1}{c}{47.59} & \multicolumn{1}{c}{50.04} & \multicolumn{1}{c}{45.55} & \multicolumn{1}{c}{33.18} & \multicolumn{1}{c}{36.47} & \multicolumn{1}{c}{29.47} & \multicolumn{1}{c}{35.24} & 37.11 \\ \midrule
\multicolumn{1}{l}{Average confidence} & \multicolumn{1}{c}{51.33} & \multicolumn{1}{c}{56.05} & \multicolumn{1}{c}{46.63} & \multicolumn{1}{c}{30.88} & \multicolumn{1}{c}{32.21} & \multicolumn{1}{c}{25.33} & \multicolumn{1}{c}{36.69} & 36.61 \\
\multicolumn{1}{l}{Box Filtering} & \multicolumn{1}{c}{50.28} & \multicolumn{1}{c}{54.43} & \multicolumn{1}{c}{45.76} & \multicolumn{1}{c}{30.50} & \multicolumn{1}{c}{31.08} & \multicolumn{1}{c}{25.32} & \multicolumn{1}{c}{34.14} & 33.72 \\
\multicolumn{1}{l}{$APLR_{box}$} & \multicolumn{1}{c}{55.94} & \multicolumn{1}{c}{59.26} & \multicolumn{1}{c}{53.04} & \multicolumn{1}{c}{41.12} & \multicolumn{1}{c}{42.99} & \multicolumn{1}{c}{35.88} & \multicolumn{1}{c}{43.83} & 46.18 \\
\bottomrule
\end{tabular}} 
\label{tab3}
\end{table*}

\begin{table}[t]
\centering
\caption{Generalizations of Omni-RES to the state-of-the-art RES models, \emph{i.e.},  LAVT~\cite{LAVT} and ReLA~\cite{GRES}. Omni-P and Omni-B denotes the use of points and boxes as the omni-labels, respectively. The proportion of fully labeled data is 10\%.}
\setlength{\tabcolsep}{5pt}
\renewcommand{\arraystretch}{1.2}
\small
\begin{tabular}{cccccccc}
\toprule
\multirow{2}{*}{Method} & \multirow{2}{*}{Model} & \multicolumn{3}{c}{\centering RefCOCO} & \multicolumn{3}{c}{RefCOCO+} \\ \cline{3-8}
& & \multicolumn{1}{c}{val} & \multicolumn{1}{c}{testA} & testB & \multicolumn{1}{c}{val} & \multicolumn{1}{c}{testA} & testB \\ \midrule
\multicolumn{1}{l}{Supervise} & \multicolumn{1}{l}{LAVT} & \multicolumn{1}{c}{57.25} & \multicolumn{1}{c}{61.63} & \multicolumn{1}{c}{52.70} & \multicolumn{1}{c}{44.24} & \multicolumn{1}{c}{49.79} & \multicolumn{1}{c}{37.04}  \\
\multicolumn{1}{l}{Omni-P} & \multicolumn{1}{l}{LAVT} & \multicolumn{1}{c}{64.70} & \multicolumn{1}{c}{68.93} & \multicolumn{1}{c}{60.27} & \multicolumn{1}{c}{50.67} & \multicolumn{1}{c}{56.73} & \multicolumn{1}{c}{42.84} \\
\multicolumn{1}{l}{Omni-B} & \multicolumn{1}{l}{LAVT} & \multicolumn{1}{c}{66.00} & \multicolumn{1}{c}{68.96} & \multicolumn{1}{c}{62.08} & \multicolumn{1}{c}{53.31} & \multicolumn{1}{c}{59.12} & \multicolumn{1}{c}{46.19}  \\ \midrule
\multicolumn{1}{l}{Supervise} & \multicolumn{1}{l}{ReLA} & \multicolumn{1}{c}{60.11} & \multicolumn{1}{c}{63.12} & \multicolumn{1}{c}{55.25} & \multicolumn{1}{c}{45.10} & \multicolumn{1}{c}{51.97} & \multicolumn{1}{c}{36.84}  \\
\multicolumn{1}{l}{Omni-P} & \multicolumn{1}{l}{ReLA} & \multicolumn{1}{c}{67.29} & \multicolumn{1}{c}{70.34} & \multicolumn{1}{c}{62.89} & \multicolumn{1}{c}{54.19} & \multicolumn{1}{c}{61.20} & \multicolumn{1}{c}{45.03} \\
\multicolumn{1}{l}{Omni-B} & \multicolumn{1}{l}{ReLA} & \multicolumn{1}{c}{68.71} & \multicolumn{1}{c}{72.08} & \multicolumn{1}{c}{65.28} & \multicolumn{1}{c}{55.86} & \multicolumn{1}{c}{62.17} & \multicolumn{1}{c}{46.72}  \\
\bottomrule
\end{tabular} 
\label{tab5}
\end{table}


\begin{table*}[t]
\centering
\caption{Comparison with the State-of-the-art methods on three RES datasets.  RefCOCO, RefCOCO+ and RefCOCOg  are used as the labeled data, and VG is used as the omni-labeled data. }
\setlength{\tabcolsep}{5pt} 
{
\small
\renewcommand{\arraystretch}{1.2}
\begin{tabular}{ccccccccccc}
\toprule
\multirow{2}{*}{Method} & \multirow{2}{*}{\begin{tabular}[c]{@{}l@{}}Visual\\ Backbone\end{tabular}} & \multirow{2}{*}{\begin{tabular}[c]{@{}l@{}}Textual\\ Encoder\end{tabular}} & \multicolumn{3}{c}{\centering RefCOCO} & \multicolumn{3}{c}{RefCOCO+} & \multicolumn{2}{c}{RefCOCOg} \\ \cline{4-11} 
& & & \multicolumn{1}{c}{val} & \multicolumn{1}{c}{testA} & testB & \multicolumn{1}{c}{val} & \multicolumn{1}{c}{testA} & testB & \multicolumn{1}{c}{val} & \multicolumn{1}{c}{test} \\ \midrule
\multicolumn{1}{c}{MAttNet~\cite{MAttNet}} & MaskRCNN-R101 & bi-LSTM & \multicolumn{1}{c}{56.51} & \multicolumn{1}{c}{62.37} & \multicolumn{1}{c}{51.70} & \multicolumn{1}{c}{46.67} & \multicolumn{1}{c}{52.39} & \multicolumn{1}{c}{40.08} & \multicolumn{1}{c}{47.64} & 48.61 \\
\multicolumn{1}{c}{CMSA~\cite{CMSA}} & DeepLab-R101 & None & \multicolumn{1}{c}{58.32} & \multicolumn{1}{c}{60.61} & \multicolumn{1}{c}{55.09} & \multicolumn{1}{c}{43.76} & \multicolumn{1}{c}{47.60} & \multicolumn{1}{c}{37.89} & \multicolumn{1}{c}{-} & - \\
\multicolumn{1}{c}{CAC~\cite{CAC}} & ResNet101 & bi-LSTM & \multicolumn{1}{c}{58.90} & \multicolumn{1}{c}{61.77} & \multicolumn{1}{c}{53.81} & \multicolumn{1}{c}{-} & \multicolumn{1}{c}{-} & \multicolumn{1}{c}{-} & \multicolumn{1}{c}{46.37} & 46.95 \\
\multicolumn{1}{c}{STEP~\cite{STEP}} & DeepLab-R101 & bi-LSTM & \multicolumn{1}{c}{60.04} & \multicolumn{1}{c}{63.46} & \multicolumn{1}{c}{57.97} & \multicolumn{1}{c}{48.19} & \multicolumn{1}{c}{52.33} & \multicolumn{1}{c}{40.41} & \multicolumn{1}{c}{-} & - \\
\multicolumn{1}{c}{BRINet~\cite{BRINet}} & DeepLab-R101 & LSTM & \multicolumn{1}{c}{60.98} & \multicolumn{1}{c}{62.99} & \multicolumn{1}{c}{59.21} & \multicolumn{1}{c}{48.17} & \multicolumn{1}{c}{52.32} & \multicolumn{1}{c}{42.11} & \multicolumn{1}{c}{-} & - \\
\multicolumn{1}{c}{CMPC~\cite{CMPC}} & DeepLab-R101 & LSTM & \multicolumn{1}{c}{61.36} & \multicolumn{1}{c}{64.53} & \multicolumn{1}{c}{59.64} & \multicolumn{1}{c}{49.56} & \multicolumn{1}{c}{53.44} & \multicolumn{1}{c}{43.23} & \multicolumn{1}{c}{-} & - \\
\multicolumn{1}{c}{LSCM~\cite{LSCM}} & DeepLab-R101 & LSTM & \multicolumn{1}{c}{61.47} & \multicolumn{1}{c}{64.99} & \multicolumn{1}{c}{59.55} & \multicolumn{1}{c}{49.34} & \multicolumn{1}{c}{53.12} & \multicolumn{1}{c}{43.50} & \multicolumn{1}{c}{-} & - \\
\multicolumn{1}{c}{CMPC+~\cite{CMPC+}} & DeepLab-R101 & LSTM & \multicolumn{1}{c}{62.47} & \multicolumn{1}{c}{65.08} & \multicolumn{1}{c}{60.82} & \multicolumn{1}{c}{50.25} & \multicolumn{1}{c}{54.04} & \multicolumn{1}{c}{43.47} & \multicolumn{1}{c}{-} & - \\
\multicolumn{1}{c}{MCN~\cite{MCN}} & Darknet53 & bi-GRU & \multicolumn{1}{c}{62.44} & \multicolumn{1}{c}{64.20} & \multicolumn{1}{c}{59.71} & \multicolumn{1}{c}{50.62} & \multicolumn{1}{c}{54.99} & \multicolumn{1}{c}{44.69} & \multicolumn{1}{c}{49.22} & 49.40 \\
\multicolumn{1}{c}{EFN~\cite{EFN}} & ResNet101 & bi-GRU & \multicolumn{1}{c}{62.76} & \multicolumn{1}{c}{65.69} & \multicolumn{1}{c}{59.67} & \multicolumn{1}{c}{51.50} & \multicolumn{1}{c}{55.24} & \multicolumn{1}{c}{43.01} & \multicolumn{1}{c}{-} & -\\
\multicolumn{1}{c}{BUSNet~\cite{BUSNet}} & DeepLab-R101 & Self-Att & \multicolumn{1}{c}{63.27} & \multicolumn{1}{c}{66.41} & \multicolumn{1}{c}{61.39} & \multicolumn{1}{c}{51.76} & \multicolumn{1}{c}{56.87} & \multicolumn{1}{c}{44.13} & \multicolumn{1}{c}{-} & - \\
\multicolumn{1}{c}{CGAN~\cite{CGAN}} & DeepLab-R101 & bi-GRU & \multicolumn{1}{c}{64.86} & \multicolumn{1}{c}{68.04} & \multicolumn{1}{c}{62.07} & \multicolumn{1}{c}{51.03} & \multicolumn{1}{c}{55.51} & \multicolumn{1}{c}{44.06} & \multicolumn{1}{c}{51.01} & 51.69 \\
\multicolumn{1}{c}{ISFP~\cite{ISFP}} & Darknet53 & bi-GRU & \multicolumn{1}{c}{65.19} & \multicolumn{1}{c}{68.45} & \multicolumn{1}{c}{62.73} & \multicolumn{1}{c}{52.70} & \multicolumn{1}{c}{56.77} & \multicolumn{1}{c}{46.39} & \multicolumn{1}{c}{52.67} & 53.00 \\
\multicolumn{1}{c}{LTS~\cite{LTS}} & Darknet53 & bi-GRU & \multicolumn{1}{c}{65.43} & \multicolumn{1}{c}{67.76} & \multicolumn{1}{c}{63.08} & \multicolumn{1}{c}{54.21} & \multicolumn{1}{c}{58.32} & \multicolumn{1}{c}{48.02} & \multicolumn{1}{c}{54.40} & 54.25 \\
\multicolumn{1}{c}{ReSTR~\cite{ReSTR}} & ViT-B & Transformer & \multicolumn{1}{c}{67.22} & \multicolumn{1}{c}{69.30} & \multicolumn{1}{c}{64.45} & \multicolumn{1}{c}{55.78} & \multicolumn{1}{c}{60.44} & \multicolumn{1}{c}{48.27} & \multicolumn{1}{c}{-} &  - \\
\multicolumn{1}{c}{MaIL~\cite{MaIL}} & ViLT & BERT & \multicolumn{1}{c}{70.13} & \multicolumn{1}{c}{71.71} & \multicolumn{1}{c}{66.92} & \multicolumn{1}{c}{62.23} & \multicolumn{1}{c}{65.92} & \multicolumn{1}{c}{56.06} & \multicolumn{1}{c}{62.45} & 62.87 \\
\multicolumn{1}{c}{CRIS~\cite{CRIS}} & CLIP-R101 & CLIP & \multicolumn{1}{c}{70.47} & \multicolumn{1}{c}{73.18} & \multicolumn{1}{c}{66.10} & \multicolumn{1}{c}{62.27} & \multicolumn{1}{c}{68.08} & \multicolumn{1}{c}{53.68} & \multicolumn{1}{c}{59.87} & 60.36 \\
\multicolumn{1}{c}{LAVT~\cite{LAVT}} & Swin-B & BERT & \multicolumn{1}{c}{72.73} & \multicolumn{1}{c}{75.82} & \multicolumn{1}{c}{68.79} & \multicolumn{1}{c}{62.14} & \multicolumn{1}{c}{68.38} & \multicolumn{1}{c}{55.10} & \multicolumn{1}{c}{61.24} & 62.09 \\
\multicolumn{1}{c}{VLT~\cite{VLT}} & Swin-B & BERT & \multicolumn{1}{c}{72.96} & \multicolumn{1}{c}{75.96} & \multicolumn{1}{c}{69.60} & \multicolumn{1}{c}{63.53} & \multicolumn{1}{c}{68.43} & \multicolumn{1}{c}{56.92} & \multicolumn{1}{c}{63.49} & 66.22 \\

\multicolumn{1}{c}{ReLA~\cite{GRES}} & Swin-B & BERT & \multicolumn{1}{c}{73.82} & \multicolumn{1}{c}{76.48} & \multicolumn{1}{c}{70.18} & \multicolumn{1}{c}{66.04} & \multicolumn{1}{c}{71.02} & \multicolumn{1}{c}{57.65} & \multicolumn{1}{c}{65.00} & 65.97 \\


\multicolumn{1}{c}{PolyFormer~\cite{polyformer}} & Swin-L & BERT & \multicolumn{1}{c}{75.96} & \multicolumn{1}{c}{78.29} & \multicolumn{1}{c}{73.25} & \multicolumn{1}{c}{69.33} & \multicolumn{1}{c}{74.56} & \multicolumn{1}{c}{61.87} & \multicolumn{1}{c}{69.20} & 70.19 \\

\midrule
\multicolumn{1}{c}{$SimRES_{omni}$ (Ours)} & Darknet53 & bi-GRU & \multicolumn{1}{c}{68.58} & \multicolumn{1}{c}{70.54} & \multicolumn{1}{c}{63.10} & \multicolumn{1}{c}{57.02} & \multicolumn{1}{c}{59.42} & \multicolumn{1}{c}{48.53} & \multicolumn{1}{c}{59.56} & 59.71 \\
\multicolumn{1}{c}{$ReLA_{omni}$ (Ours)} & Swin-B & BERT & \multicolumn{1}{c}{\textbf{78.69}} & \multicolumn{1}{c}{\textbf{80.66}} & \multicolumn{1}{c}{\textbf{75.53}} & \multicolumn{1}{c}{\textbf{71.06}} & \multicolumn{1}{c}{\textbf{74.60}} & \multicolumn{1}{c}{\textbf{63.72}} & \multicolumn{1}{c}{\textbf{71.40}} & \textbf{73.19} \\
\bottomrule
\end{tabular}}
\label{tab_sota}
\end{table*}

\subsection{Implementation Details}
The model configurations, such as visual backbone, are kept the same with previous works~\cite{SimREC}.
Adam is used as the optimizer with a learning rate of 1$e$-4, and a batch size of 64. In Omni-RES, EMA~\cite{EMA} coefficient is set to 0.9996 for updating the Teacher model.
By default, the thresholds applied for filtering the teacher's prediction are set between 0.7 and 0.2, and $\gamma$ are set to 0.2. For omni-training with boxes and points, the threshold  is adjusted between 0.5 and 0.2, and the weight of omni-loss is set to 1. More details can refer to our source code.

\subsection{Quantitative Analysis}

\textbf{Ablation Study.} We first ablate Omni-RES with different types of omni-labels and compare it with \emph{the semi- and fully supervised baseline} in Tab.~\ref{tab2}. 
Omni-None also refers to the semi-supervised baseline. 
From this table, we first observe that the semi-supervised baseline (Omni-None) can already outperform fully supervised baseline by a large margin, \emph{e.g.,} +9.52\% on 1\% RefCOCO. With the help of our omni-training, \emph{e.g.,} Omni-Point and Omni-Box, the performance boost significantly, \emph{e.g.,} +24.47 by Omni-Box on RefCOCO \textit{testA}. However, we also see that the performance of Omni-Point varies greatly under different splits of labeled data.
With 10\% labeled data, its performance is close to Omni-Box, while the gap becomes larger when the label information is less,~\emph{e.g.}, 1\%.
To explain, when the label information is very limited, the quality of pseudo-masks are too poor to meet the referring point, \emph{e.g.}, very scattered segmentation results.
This case also reflects the challenge of RES towards omni-supervised learning and the inferior pseudo-label quality than other tasks.

In Tab.~\ref{tab3}, we compare the proposed~\emph{Active Pseudo-Label Refinement} (APLR) with a set of alternative filtering schemes. 
In particular, $APLR_{point}$ and $APLR_{w/o filtering}$ denote the used metric in  Eq.~\ref{equ:sel_point} and the the removal of filtering strategy, respectively.
Point distance and $APLR_{box}$ denote the filtering metric defined by geometry distance and the metric defined in Eq.~\ref{equ:sel_box}, respectively.
Compared with it, \emph{box filtering} only suppress the pixels outside the box.
~\emph{Average confidence} means the average positive pixel confidence within the omni-box. 
From Tab.~\ref{tab3}, we can see that the alternative filtering strategies can improve the omni-RES performance to some extent. However, their benefits are very marginal. Meanwhile, it is also not very significant to directly apply the filtering strategy from other Omni-methods~\cite{Omni-DETR},~\emph{i.e.} point distance. 
The proposed APLR for omni-point and omni-box show obvious advantages to the alternative methods, especially for~\emph{$APLR_{w/o filtering}$} and~\emph{box filtering}.
These results on one hand confirm the effectiveness of APLR for Omni-RES, and on the other hand suggest that the significance of pseudo-label quality.

\textbf{Generalization to LAVT and ReLA.} To examine the generalization ability of Omni-RES, we further apply it to two advanced RES models, namely LAVT~\cite{LAVT} and ReLA~\cite{GRES}, of which results are given in Tab.~\ref{tab5}. 
In particular, LAVT and ReLA demonstrate strong supervised performance on 10\% RefCOCO.
Even so, we can still observe the significant gains of Omni-RES on these methods. For instance, the relative performance improvements of Omni-Point and Omni-Box can be up to +5.80\% and +9.15\% on RefCOCO+~\textit{testB}, respectively. For ReLA, the improvements are more significant, \emph{e.g.,} +8.19\% and +9.88\%.
These results well confirm that Omni-RES can collaborate well with existing SOTA RES models, and further improve their performance.

~\textbf{Comparison with the State-of-the-art.} 
In Tab.~\ref{tab_sota}, we compare SimRES and ReLA trained by Omni-Box with the existing RES approaches. In particular, we conduct the training of Omni-Box with the labeled data of three RefCOCO-series datasets and the weakly labeled data of Visual Genome~\cite{VG}\footnote{ the images in the validation and test splits of three benchmarks are removed.}. From this table, the first observation is that Omni-Box obviously outperform its fully supervised counterpart,~\emph{e.g.,} +7.22\% on RefCOCOg~\textit{test}.
With the help of Omni-Box, $ReLA_{omni}$ can achieve new SOTA performances on three RES datasets, \emph{e.g.,} 80.66\% on RefCOCO \textit{testA}.
When compared to the existing SOTA model, ~\emph{i.e.,} PolyFormer~\cite{polyformer}, the performance gains of $ReLA_{omni}$ become significant on the more challenging datasets, such as RefCOCO+ \textit{testB}, and RefCOCOg \textit{test}. These results also suggest that our Omni-training can facilitate the learning of these difficult examples. Overall, these results further confirm the effectiveness of Omni-RES, and also validate our motivation about the use of weakly labeled data.

\subsection{Qualitative Analysis}

\begin{figure}[!htb]
\centering
\includegraphics[width=0.48\textwidth]{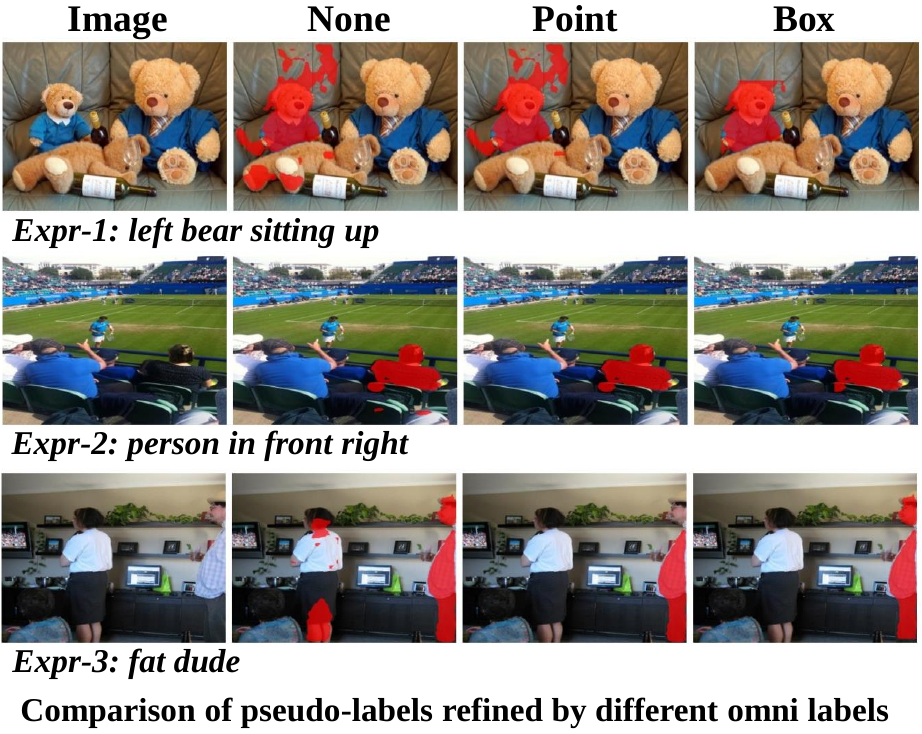}
\caption{Visualizations of Omni-RES. Without any refinements, the pseudo-masks are often scattered and erroneous, while Omni-RES can use the proposed Active Pseudo-Label Refinement to improve the quality of pseudo-masks based on omni-labels of points or boxes.}
\label{fig:vis}
\end{figure}

To obtain deep insights into Omni-RES, we further visualize their results of pseudo-label refinements in Fig.~\ref{fig:vis}, which shows the pseudo-masks with and without the proposed APRL. 
We can first observe that the qualities of original pseudo-labels are poor, which often fails to completely mask the referent or has incorrect predictions. 
For instance, the model segment several parts of two people in the 3-th example.
With APRL,~\emph{i.e.}, Point and Box, their qualities are greatly improved. 
In terms of APRL with the omni-labels of points, the segmentations for the incorrect targets can be well handled,~\emph{e.g.}, the 2-th examples, suggesting its merits in locating referents. 
However, its refinement to the mask contour is still insufficient.
For example, it still includes the background to the referent of bear in the 1-st example.
In contrast, APRL with grounding boxes can filter out all incorrect mask pixels outside the box, leading to better masks in details. Meanwhile, its ability of spatial grounding is also as well as the points. Overall, these results further confirm the merits of Omni-RES.


\section{Conclusions}
In this paper, we focus on the issue of expensive instance-level annotation and insufficient label information of RES, and propose a novel omni-supervised learning task, termed Omni-RES. Omni-RES aims to make full use of unlabeled, fully-labeled and weak labels to facilitate efficient RES training,~\emph{i.e.}, the grounding boxes and referring points. To accomplish this task, we also propose a strong basline method based on the recently popular teacher-student framework. In this method, the omni-labels, i.e., such as referring points and grounding boxes are not directly transformed into supervision signals for RES training. Instead, we use them as reference to produce high-quality pseudo-masks for teacher-student learning. 
To validate Omni-RES, we apply it a simple base model called SimRES and conduct extensive experiments on three RES benchmarks. 
The experimental results show the great advantages of Omni-RES than the supervised and semi-supervised baselines. Notably, with only 10\% fully labeled data, Omni-RES can achieve the 100\%s supervised performance, strongly confirming the benefits of Omni-RES. In addition, the generalization of Omni-RES is also validated on recently proposed RES models called LAVT and ReLA.
Finally, we achieve the SOTA result using ReLA model with extra Omni-Box labels, which obviously outperforms the existing SOTA model called PolyFormer.

\section{Acknowledgements}
This work was supported by National Key R\&D Program of China (No.2022ZD0118201) , the National Science Fund for Distinguished Young Scholars (No.62025603), the National Natural Science Foundation of China (No.U21B2037, No.U22B2051, No.62176222, No.62176223, No.62176226, No.62072386, No.62072387, No.62072389, No.62002305 and No.62272401), and the Natural Science Foundation of Fujian Province of China (No.2021J01002,  No.2022J06001), and the China Fundamental Research Funds for the Central Universities (Grant No.20720220068).

\bibliography{aaai24}

\end{document}